%% file: bmvc_final.tex
\documentclass{bmvc2k}

\title{Signing Outside the Studio: 

Benchmarking Background Robustness for Continuous Sign Language Recognition}

\addauthor{Youngjoon Jang}{wgs01088@kaist.ac.kr}{1}
\addauthor{Youngtaek Oh}{youngtaek.oh@kaist.ac.kr}{1}
\addauthor{Jae Won Cho}{chojw@kaist.ac.kr}{1}
\addauthor{Dong-Jin Kim}{djdkim@hanyang.ac.kr}{2}
\addauthor{Joon Son Chung}{joonson@kaist.ac.kr}{1}
\addauthor{In So Kweon}{iskweon77@kaist.ac.kr}{1}

\addinstitution{
 Korea Advanced Institute of Science
and Technology (KAIST)\\
 Daejeon, Republic of Korea\\
}

\addinstitution{
 Hanyang University\\
 Seoul, Republic of Korea
}

\runninghead{Jang et al.}{Signing Outside the Studio}

\usepackage{graphicx}
\usepackage{amsmath}
\usepackage{amssymb}
\usepackage{booktabs}
\usepackage{extra_styles}
\usepackage{subfigure}
\usepackage[capitalize]{cleveref}

\usepackage{algorithm}
\usepackage{algorithmic}
\usepackage[normalem]{ulem}

\crefname{section}{Sec.}{Secs.}  
\Crefname{section}{Section}{Sections}
\Crefname{table}{Table}{Tables}
\crefname{table}{Tab.}{Tabs.}
\crefname{algorithm}{Alg.}{Algs.}  
\Crefname{algorithm}{Algorithm}{Algorithms}

\begin{document}
\maketitle
\input{sec/0_abstract}

\input{sec/1_introduction}
\input{sec/2_related}
\input{sec/2.1_benchmark}
\input{sec/3_method}

\input{sec/4_results}

\input{sec/5_conclusions}

\input{sec/6_ack}

\bibliography{egbib}
\end{document}


\maketitle
\appendix

\input{sec/6_supplementary}

\bibliography{egbib}

%% file: sec/0_abstract.tex
\begin{abstract}
The goal of this work is background-robust continuous sign language recognition. Most existing Continuous Sign Language Recognition (CSLR) benchmarks have fixed backgrounds and are filmed in studios with a static monochromatic background. However, signing is not limited only to studios in the real world. In order to analyze the robustness of CSLR models under background shifts, we first evaluate existing state-of-the-art CSLR models on diverse backgrounds. To synthesize the sign videos with a variety of backgrounds, we propose a pipeline to automatically generate a benchmark dataset utilizing existing CSLR benchmarks. Our newly constructed benchmark dataset consists of diverse scenes to simulate a real-world environment. We observe even the most recent CSLR method cannot recognize glosses well on our new dataset with changed backgrounds. In this regard, we also propose a simple yet effective training scheme including (1) background randomization and (2) feature disentanglement for CSLR models. The experimental results on our dataset demonstrate that our method generalizes well to other unseen background data with minimal additional training images.
Our dataset is available \href{https://github.com/art-jang/Signing-Outside-the-Studio}{here}.
\end{abstract}

%% file: sec/1_introduction.tex
\begin{figure}[t]
    \centering
        \subfigure[][Grad-CAM~\cite{selvaraju2017grad} activation maps.]{
            \includegraphics[width=.40\columnwidth]{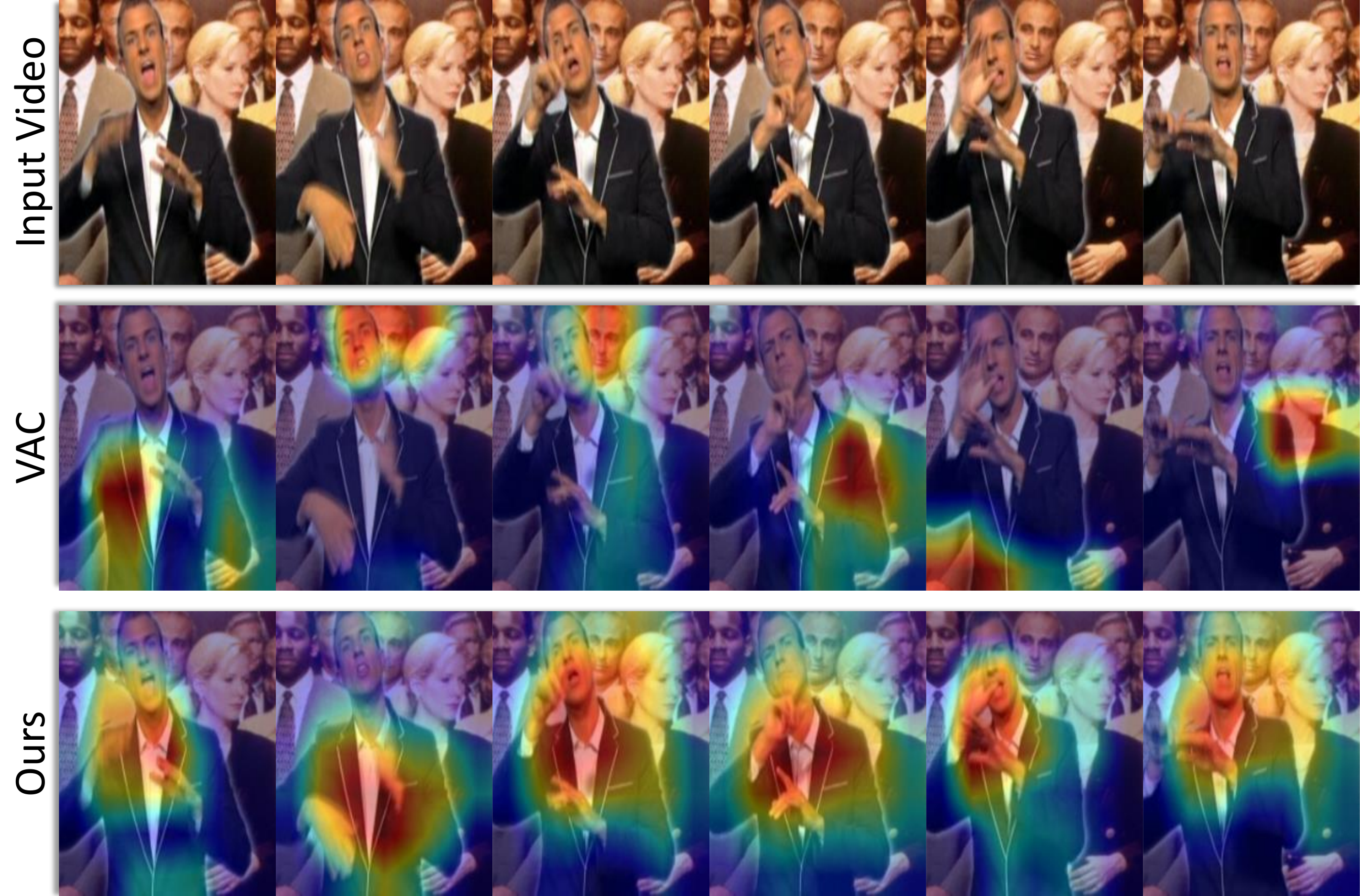}
            \label{fig:teaser}
        }\quad \quad
        \subfigure[][WER scores from test benchmarks.]{
            \includegraphics[width=.40\columnwidth]{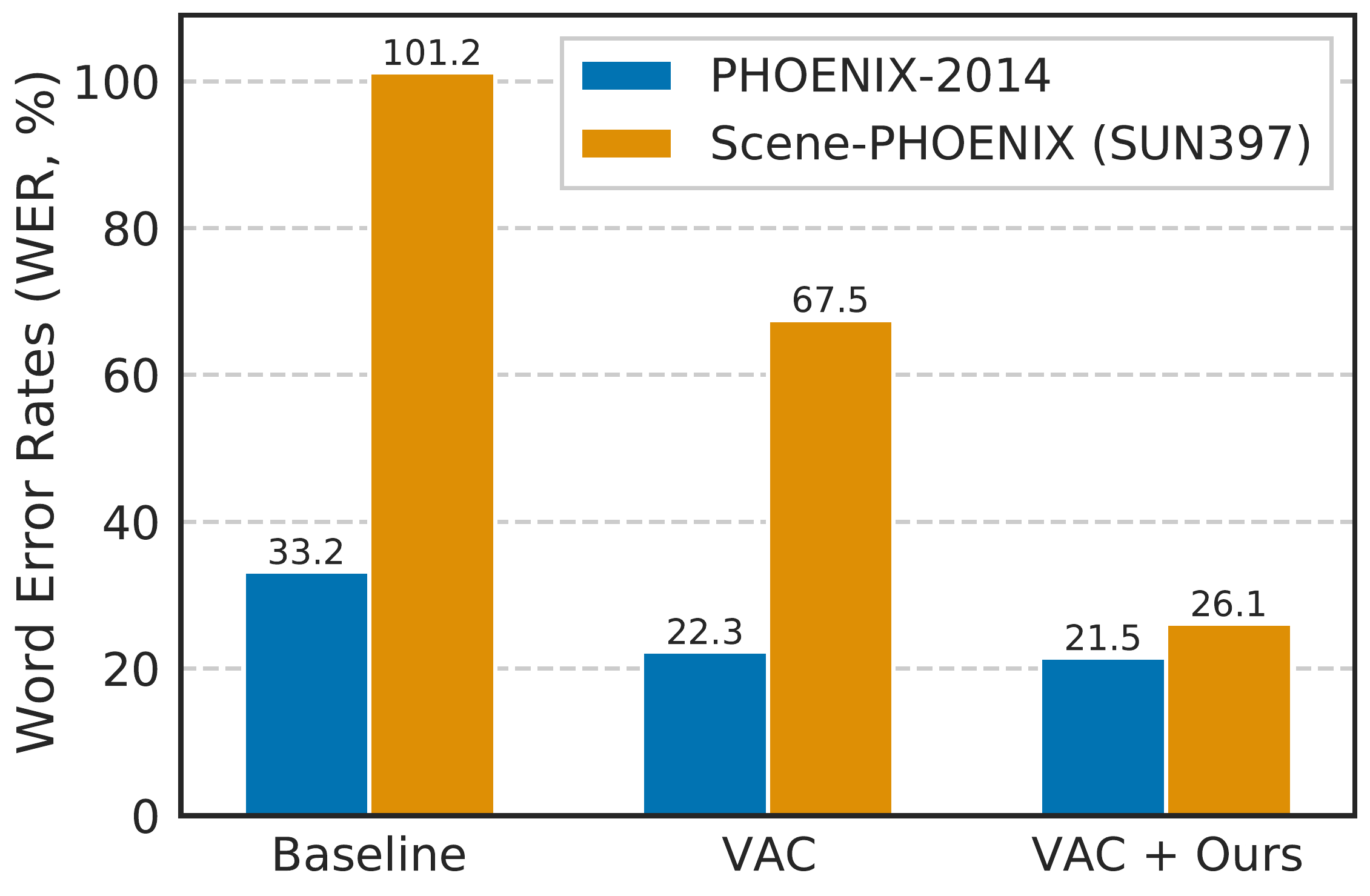}
            \label{fig:eval_background}
        }
    \caption{
        We evaluate CSLR models trained on PHOENIX-2014~\cite{koller2015continuous} under our newly proposed Scene-PHOENIX benchmark.
        \textbf{(a)} VAC trained on monochromatic background sign videos fails to attent to the signer in the video and attends to other people in the background.
        \textbf{(b)} Both Baseline (ResNet18~\cite{he2016deep} + LSTM) and VAC~\cite{min2021visual} severely degrade when tested on Scene-PHOENIX. 
        In contrast, our framework can still capture signer's dominant expressions and favorably close the gap between test splits of the original PHOENIX-2014 and Scene-PHOENIX. We further detail experimental setup in~\cref{sec:experiments}.
    }
\label{fig:teaser_total}
\end{figure}

\section{Introduction}
\label{sec:intro}
Most publicly available CSLR benchmarks are curated from either studio or TV broadcasts, where background images are fixed and monochromatic~\cite{huang2018video,koller2015continuous,2021ksl}. 
In a deployment scenario, 
these backgrounds are dissimilar to situations where real world communications occur,
potentially limiting the practicality of CSLR models.
A na\"ive solution to this would be constructing a new dataset outside the studio, but the cost of extensive gloss annotations as well as collecting sign videos with skilled signers present significant challenges. 

To tackle this issue, we first propose an automatic pipeline for CSLR benchmark dataset generation that re-uses existing CSLR datasets to synthesize a new dataset with various backgrounds for evaluating the robustness of background changes with minimal human intervention.
During this process, we select natural scene images from scene datasets~\cite{yu2015lsun,xiao2016sun} and carefully incorporate them on the test set of existing benchmark with a 
person mask. 
We make variants of development and test splits of PHOENIX-2014~\cite{koller2015continuous} with our automated pipeline and name our benchmark dataset with diverse backgrounds \emph{Scene-PHOENIX}. 

Based on our Scene-PHOENIX dataset, we find that current CSLR approaches are not robust to background shifts.
For example, we observe that 
VAC~\cite{min2021visual}, a state-of-the-art CSLR method, severely degrades on our new evaluation criteria.
Hence, we find the need for addressing background shift issue to improve practicality.
To this end, we further propose a simple yet effective training scheme.
First, we employ background randomization, where a sign video for training is combined with a scene image via mixup~\cite{zhang2017mixup} to simulate background shifts. 
Then, 
we design a Disentangling Auto-Encoder (DAE), which aims to disentangle the signer feature and the background feature in the latent space. 
We emphasize that we use only a few background images during training, and 
the DAE can disentangle the signers from the background without additional annotations (\eg, body keypoint, person mask) as shown in~\cref{fig:teaser}. 
From the experimental results based on Scene-PHOENIX in~\cref{fig:eval_background}, we show that our method greatly reduces the gap between performance on test set with diverse backgrounds and with monochromatic backgrounds while boosting the performance on the original test set as well.

Our contributions are as follows:
(1) To the best of our knowledge, we are the first to study the background shift issue in CSLR. We propose an automatic benchmark dataset generation pipeline that can be applied to any CSLR dataset and generate our dataset, Scene-PHOENIX.
(2) We propose a new training scheme for CSLR, including background randomization and Disentangling Auto-Encoder (DAE) to improve robustness to background shifts. Our approach can be readily integrated to any CSLR models using only a small number of extra images, without any annotations.
(3) We experimentally show even the recent state-of-the-art CSLR models suffer significant performance degradation on Scene-PHOENIX. We also show that our approach effectively improves the robustness to background shifts while maintaining the performance from the original test data.

%% file: sec/2_related.tex
\section{Related Work}
\label{sec:related}
\noindent \textbf{Background Bias in Sign Videos.}
The Continuous Sign Language Recognition (CSLR) task~\cite{koller2016deepA,camgoz2017subunets,pu2018dilated,huang2018video} aims to predict a sequence of glosses from a sign video.
To efficiently construct benchmarks for the CSLR task on a large-scale, sign videos are often crawled from TV programs~\cite{koller2015continuous,albanie2020bsl,Albanie2021bobsl}.
Sign videos are also captured in lab environments, to obtain multi-view videos~\cite{ko2019neural,duarte2021how2sign} or 
body pose and depth~\cite{von2007towards,huang2018video,duarte2021how2sign}. 
Based on such benchmarks, 
numerous CSLR studies have been proposed~\cite{yang2019sf,cheng2020fully,niu2020stochastic,min2021visual,hao2021self,camgoz2017subunets,huang2018video,koller2019weakly,zhou2020spatial}.
In isolated SLR tasks, \citet{carneiro2021efficient} leverage background replacement for data augmentation at train-time.
However, robustness to \emph{background shift} has not been explored in the CSLR field~\cite{huang2018video,koller2015continuous,cihan2018neural}.
As most CSLR benchmarks have monochromatic backgrounds~\cite{huang2018video,koller2015continuous,cihan2018neural}, we observe that models are biased toward the background of the training data and cannot be generalized in videos with diverse background distributions (\eg, indoor or outdoor scenes common in daily life).
In this regard, we release a new benchmark dataset, named Scene-PHOENIX, to measure the robustness of CSLR models to background shifts without the need for collecting the new sign videos.

\noindent \textbf{Robustness to Distribution Shift.}
Addressing the distribution shift is a crucial research problem since deep learning models are fragile to testing distribution different from the training~\cite{shen2021towards}. In this aspect, various benchmarks have been proposed to measure the robustness under distribution shifts~\cite{hendrycks2018benchmarking,recht2019imagenet,engstrom2019exploring,Hendrycks_2021_CVPR,hendrycks2021many,senocak2022audio,cho2022generative,oh2022daso,kim2018disjoint,kim2019image}, and this problem has been extensively studied in broad research fields~\cite{ganin2016domain,bousmalis2017unsupervised,wang2021tent,shin2021labor,li2017deeper,li2018learning,zhou2021domain,hendrycks2020augmix,Li_2021_CVPR,geirhos2018imagenettrained,choi2019can,barbu2019objectnet,nam2021reducing}.
Among them, benchmarking robustness~\cite{hendrycks2018benchmarking} and resolving scene bias~\cite{choi2019can,mo2021object} or distribution shift~\cite{yue2019domain,nam2021reducing} are the most related to our problem setup.
Different from the aforementioned works, we first explore the background shift issue in the CSLR task with a newly synthesized benchmark. We then propose a simple and versatile training scheme to train a background-agnostic CSLR model while preserving performance in the original test data without background change.

%% file: sec/2.1_benchmark.tex
\begin{figure}[t]
    \centering
    \includegraphics[width=.5\columnwidth]{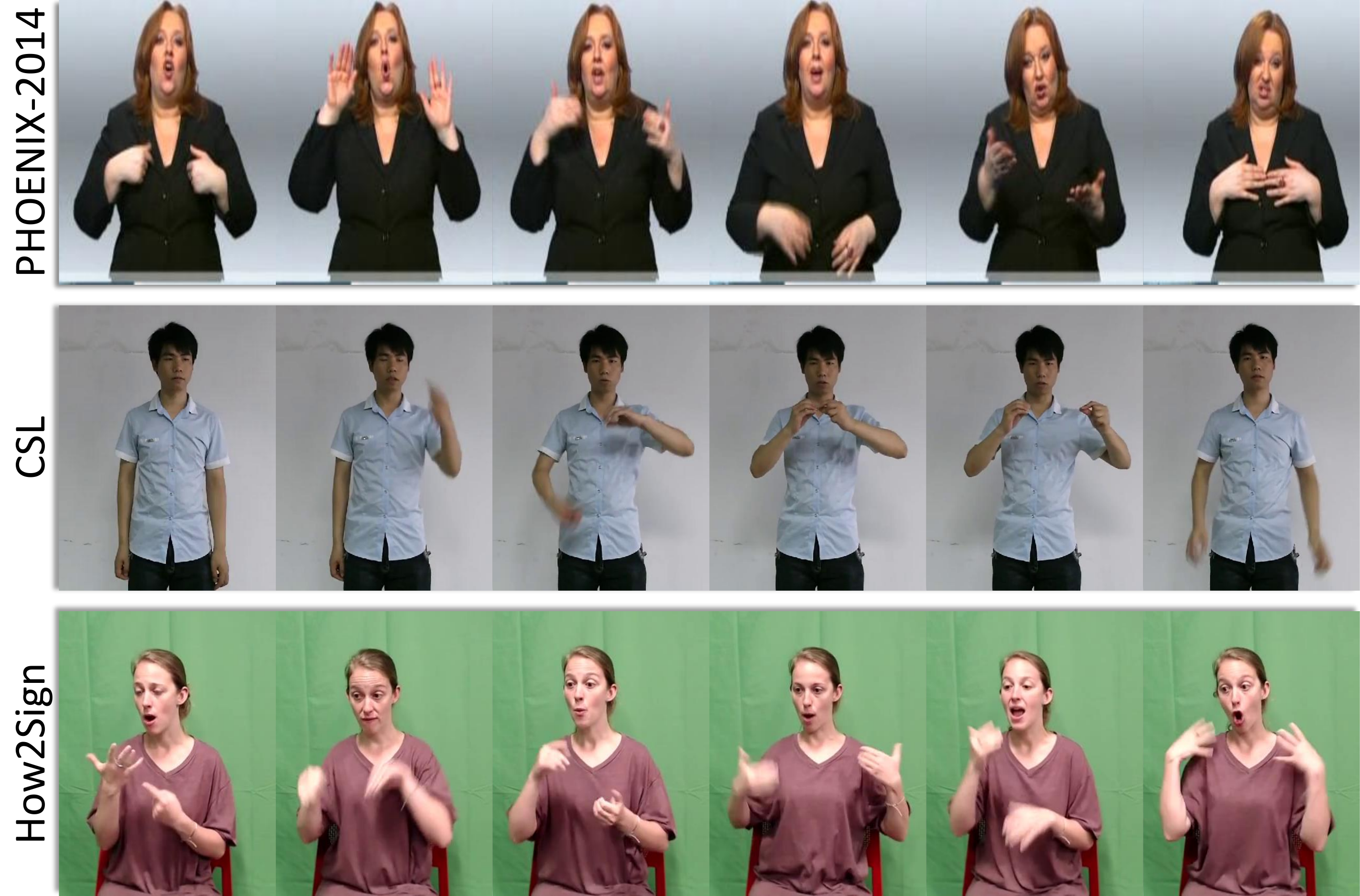}
     \caption{
     Comparison of videos in Continuous Sign Language Recognition (CSLR) dataset~\cite{koller2015continuous,cihan2018neural,duarte2021how2sign}. All videos display monochromatic backgrounds.
     }
    \label{fig:sign_dataset_examples}
\end{figure}

\section{Benchmarking Background Robustness}
\label{sec:benchmark}
\noindent\textbf{Background Bias.}
Continuous Sign Language Recognition (CSLR) videos are collected from weather forecasts~\cite{koller2015continuous} or studio recordings~\cite{huang2018video,duarte2021how2sign}, causing backgrounds to be fixed and monochromatic as in~\cref{fig:sign_dataset_examples}. As a result, CSLR models are trained and evaluated on a dataset with the same background distribution.
However, the model is biased toward the background of the training data and fails to generalize in sign videos with \emph{unseen} backgrounds.
To support the hypothesis, we train CSLR models on PHOENIX-2014~\cite{koller2015continuous}, and compare the Word Error Rate (WER)\footnote{WER = ({\#substitutions} + \text{\#deletions} + \text{\#insertions}) / ({\text{\#words in reference}})} scores~\cite{koller2015continuous} between the original test split and our new test split, \emph{Scene-PHOENIX}.
As shown in~\cref{fig:eval_background}, the performance of the baseline CSLR model \emph{completely} degrades on Scene-PHOENIX (WER: 33.2\% $\rightarrow$ 101.2\%).
Even VAC~\cite{min2021visual} degrades in performance (WER: 22.3\% $\rightarrow$ 67.5\%). 
Furthermore, feature activations in~\cref{fig:teaser} show that VAC is unable to consistently attend to the signer. On the other hand, our method significantly improves WER on background shifts, further closing the gap from the original test split. 

\noindent\textbf{Robust Dataset Construction.} 
In order to tackle the severe performance degradation of CSLR models during background shifts, we propose a new CSLR benchmark dataset for background shift evaluation. As collecting and annotating new sign language videos are expensive and time-consuming, we propose an effective alternative, an automatic CSLR benchmark dataset generation algorithm that utilizes existing datasets as shown in \cref{fig:test_gen}.

For a given sign video in PHOENIX-2014, a set of person mask is obtained for every frame by employing an open-source pretrained semantic segmentation network\footnote{\url{https://github.com/thuyngch/Human-Segmentation-PyTorch}}. 
Then, we apply background matting with the masks and replace the background with external natural scene images across all frames.
We utilize LSUN~\cite{yu2015lsun} and SUN397~\cite{xiao2016sun}, which contain a wide range of indoor and outdoor scenes of 10 and 397 classes respectively, as our background images.
Note that we uniformly distribute each scene class across the sign samples for the test split of PHOENIX-2014 with 629 samples, 62-63 videos are assigned for each class of LSUN, and 1-2 videos are assigned for each class of  SUN397.
Each synthesized set is called a \emph{Split} and we generate three \emph{Splits} for each dataset (\ie,~LSUN and SUN397) for reliable evaluation.
Note that this method can be applied to other CSLR datasets to create additional benchmarks.

%% file: sec/3_method.tex
\section{Background Agnostic Framework}
\label{sec:method}


We design a new framework that enables models to learn CSLR in a background agnostic manner
as shown in~\cref{fig:architecture}. Our framework comprises of (1) \emph{Background Randomization} (BR), which simply generates a sign video with new background via mixup~\cite{zhang2017mixup} to simulate background shift,
and (2) \emph{Disentangling Auto-Encoder} (DAE) that aims to disentangle the signer from videos with background in latent space obtained by the query encoder $e^q$ and the key encoder $e^k$.
The Teacher Network from \Fref{fig:architecture} serves as the target network, taking the original video and then provides positive and negative pairs from spatially encoded features, inspired by~\cite{he2020momentum,grill2020bootstrap}.
Finally, the sequence model is provided with the disentangled signer component of latent feature $h^q_s$.
We adapt the meta-architecture (CNN+LSTM) used in~\cite{zhou2020spatial,cheng2020fully,min2021visual,hao2021self} as our~\emph{Baseline} and train this model via CTC loss~\cite{graves2006connectionist}.

\begin{figure*}[!t]
    \centering
        \subfigure[][Generation of Scene-PHOENIX.]{
        \includegraphics[width=.45\linewidth]{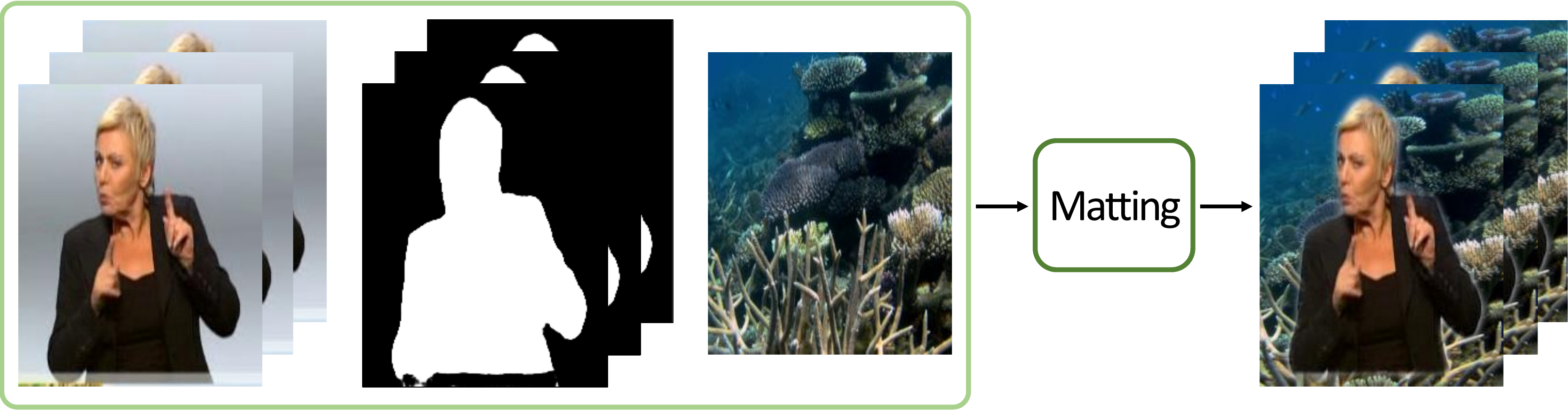}
        \label{fig:test_gen}}\quad\quad
        
        \subfigure[][Background randomization \emph{for training}.
        ]{
        \includegraphics[width=.45\linewidth]{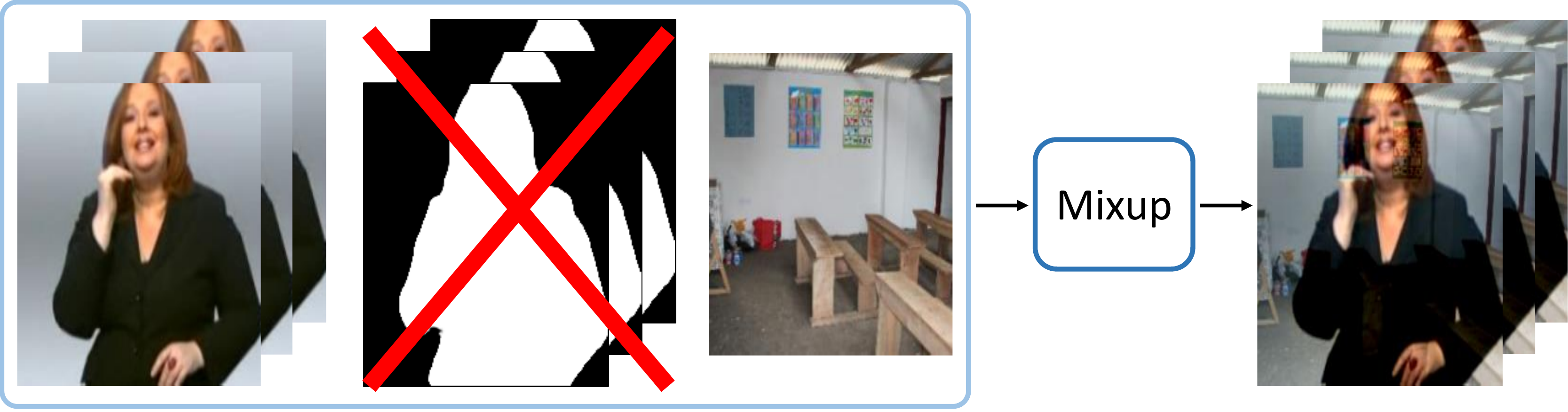}
        \label{fig:dr_gen}}

\caption{Data generation. {\textbf{(a)}} For Scene-PHOENIX, background matting is performed with scene images using person masks. {\textbf{(b)}} For training set, we apply mixup~\cite{zhang2017mixup} between a sign video and a scene image \emph{without person masks} to reduce additional labeling cost in training.}
\label{fig:data_geneartion}
\vspace{-2mm}
\end{figure*}

\subsection{Background Randomization}
\label{sec:dom_rand}
We propose to leverage additional natural scene images to create videos with randomized backgrounds to bridge the gap from real world shifts.
However, adding scenes used at test time during training can be seen as a \emph{trivial} way to enhance robustness.
Thus, to better test robustness and reduce potential costs, we limit the number of scene images available during training.
In detail, we denote a variable $K$, where $K$ is the number of images that we sample per class within the LSUN dataset. For example, if $K=1$, we select 1 image per class, and LSUN has 10 classes, resulting in a total of 10 images for BR.
Then, as shown in~\cref{fig:dr_gen}, we obtain a convex sum~\cite{zhang2017mixup} of the target video with random background images.
While \citet{carneiro2021efficient} use person masks for changing backgrounds in training ISLR models, we emphasize BR is done without masks to reduce additional labor costs.

\subsection{Disentangling Auto-Encoder}
\label{sec:dae}
While Background Randomization (BR) improves a CSLR model's background shift robustness, we additionally propose a light-weight Disentangling Auto-Encoder (DAE) that further improves this robustness. In designing DAE, we hypothesize that the input video can be separated into a signer feature and a background feature in the embedding space.

\begin{figure*}[t]
    \centering
    \includegraphics[width=.90\linewidth]{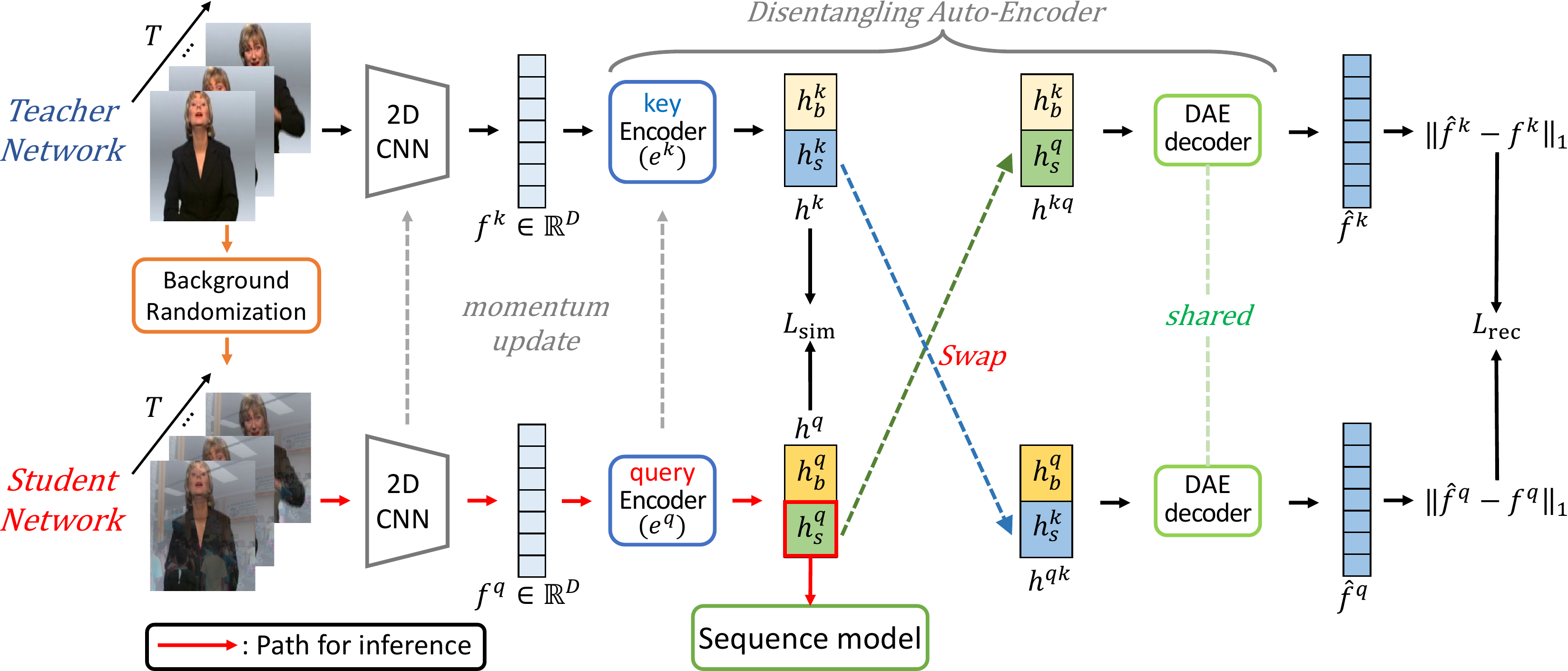}
    \caption{The overall architecture of the proposed model. The original video passes through Teacher Network, and the background-randomized video passes through Student Notwork.
    In the latent space, the signer features $(h_s^k, h_s^q)$ are swapped with each other.
    Then, the swapped features $(h^{kq}, h^{qk})$ are input to the shared DAE decoder for reconstructing the original features $f^k$ and $f^q$.
    Note the Red arrows show the path during inference.}
    \label{fig:architecture}
    \vspace{-2mm}
\end{figure*}

As shown in~\Fref{fig:architecture}, our framework consists of a \emph{Teacher Network} and a \emph{Student Network}~\cite{hinton2015distilling,he2020momentum,kim2020detecting,kim2021acp++,cho2021dealing} that both have the same network architecture. The teacher network takes the original sign videos, and the student network takes the background-randomized sign videos as their inputs. Each input videos pass through a spatial feature extractor (2D CNN) and then are flattened without average pooling to obtain $D$ dimensional vectors $f^k$ and $f^q$. Sequentially, key and query encoders embed $f^k$ and $f^q$ into $h^k$ and $h^q$ respectively. Here, we physically divide each latent feature ($h^k, h^q)$ into two parts and we assume that the divided latent features consist of signer feature $h_s$ and background feature $h_b$.

In order to embed more discriminative latent features, we give an additional training objective so that the signer features (\eg, $h^q_s$ and $h^k_s$) should pull each other and the background features (\eg, $h^q_b$ and $h^k_b$) push against each other.
We employ the cosine similarity losses $L^{pos}_{\text{sim}}$ and $L^{neg}_{\text{sim}}$ for pulling and pushing respectively. The similarity loss is formulated as follows:
\begin{align}
    L^{pos}_{\text{sim}}(x_1,x_2) = 1 - \text{cos}(x_1, x_2), \text{\quad}
    L^{neg}_{\text{sim}}(x_1,x_2) = \max(0, \text{cos}(x_1, x_2) - \Delta),
\end{align}
where $\Delta$ is a hyperparameter for the margin, which penalizes based on its value.
The final similarity loss is given by:
\begin{equation}
    L_{\text{sim}}=L^{pos}_{\text{sim}}(h^q_s, h^k_s) + L^{neg}_{\text{sim}}(h^q_b, h^k_b).
\end{equation}

In the case that the latent features $(h^q, h^k)$ are perfectly disentangled into signer feature $(h_s)$ and background feature $(h_b)$, there should be no difference between the signer features $h^q_s$ and $h^k_s$. To try to enforce this, we first \emph{swap} the signer features of the teacher network and the student network. $h^{kq}$ and $h^{qk}$ denote the swapped features. Here, we train our DAE decoder that re-projects the latent features $(h^q, h^k)$ back into the original feature space after the 2D CNN by reconstructing $f^q$ and $f^k$ from the swapped features $h^{kq}$ and $h^{qk}$.
To enforce such reconstruction, $L_{\text{rec}}$ is measured as L1 distances of respective reconstructed features:
\begin{equation}
    L_{\text{rec}}=\lVert \hat{f}^q - f^q\rVert_{1} + \lVert \hat{f}^k - f^k\rVert_{1},
\end{equation}
where $\hat{f}^q$ and $\hat{f}^k$ are re-projected features from the DAE decoder. 

Finally, only $h^q_s$ is passed through the sequence model to predict gloss sequences by propagating CTC loss~\cite{graves2006connectionist}, so the network can focus more on the signer in a background agnostic manner.
During inference, the teacher network and DAE decoder are discarded, causing the inference overhead to be negligible.
Note that the teacher network is updated by momentum update~\cite{he2020momentum}.

\subsection{Objective Function.}
The final objective of Ours when integrated with VAC~\cite{min2021visual} network is as follows:
\begin{equation}
    L_{\text{total}} = \underbrace{L_{\text{CTC}} + L_{\text{VE}} + \alpha L_{\text{VA}}}_{L_{\text{VAC}}} + \underbrace{L_{\text{sim}} + L_{\text{rec}}}_{L_{\text{DAE}}},
\end{equation}
where the first three terms correspond to VAC, and the final two terms are used for train the DAE. 
In our framework, we empirically find that lower value of $\alpha$ accelerates the convergence of the model as a whole and set $\alpha$ to 3.

%% file: sec/4_results.tex
\section{Experiments}
\label{sec:experiments}
\noindent \textbf{Dataset.} 
We utilize the publicly available PHOENIX-2014~\cite{koller2015continuous} dataset to validate our framework.
In order to evaluate the robustness of models to background changes, we construct Scene-PHOENIX by synthesizing the background and perform various experiments on both PHOENIX-2014 and Scene-PHOENIX.
To generate Scene-PHOENIX, we use two large-scale scene datasets, LSUN~\cite{yu2015lsun} and SUN397~\cite{xiao2016sun}, which consist of about 1M images with 10 classes and about 0.1M images with 397 classes, respectively.

\noindent \textbf{Evaluation Protocol.}
We measure the performance of CSLR models by WER score.
Scene-PHOENIX consist of one Dev \emph{Split}, and three Test \emph{Splits} respectively on both LSUN and SUN397.
We report the average of WERs from all Test \emph{Splits}.
$\text{WER}^{\text{LSUN}}$ and $\text{WER}^{\text{SUN}}$ denote WER in the test set in which Scene-PHOENIX video background is randomized with LSUN and SUN397 respectively.

\noindent \textbf{Implementation details.}
The proposed Disentangling Auto-Encoder (DAE) has an encoder-decoder architecture, and both the encoder and the decoder consists of two fully-connected layers. The input video frames are first resized to 256$\times$256, followed by random crop to 224$\times$224 at the same location of all the frames and random horizontal flip with 50\% probability. We then randomly insert duplicated frames up to 20\% in total length, followed by random deletion of frames up to 20\% of the whole length. 
We train the network using Adam optimizer~\cite{kingma2014adam} with batch size 2 and weight decay $10^{-4}$ for 100 epochs. 
The initial learning rate is $10^{-4}$ and the cosine scheduler is adopted to decay the rate.
Our framework is implemented using PyTorch~\cite{paszke2019pytorch}.

\input{tab/main_table}

\subsection{Main Results}

\input{tab/abl_br}

\noindent\textbf{VAC-Oracle.}
We first introduce a VAC~\cite{min2021visual} based Oracle as a frame of reference. The Oracle set uses all of the images within the LSUN dataset (over 1$M$ images), and instead of using our simple BR method, we generate all images in the Oracle set using background matting as illustrated in~\Cref{fig:test_gen}, identical to the Scene-PHOENIX set. Then, we report the VAC model trained on this Oracle set as VAC-Oracle in \Cref{tab:main_table}.

\noindent \textbf{Baseline.} \Cref{tab:main_table} shows the comparison of our framework including BR and DAE with other methods. The Baseline model gives the WER scores larger than 100\% in all splits of Scene-PHOENIX. While using pretrained feature extractor on ImageNet~\cite{deng2009imagenet} (w/ pretrain) can be beneficial for the background shifts ($WER^{SUN}_{Test}$: 101.2\% $\rightarrow$ 72.7\%), there still exists large performance gap between PHOENIX-2014 and Scene-PHOENIX. In contrast, applying our BR and DAE to the pretrained Baseline dramatically improves the performance on Scene-PHOENIX ($WER^{LSUN}_{Test}$: 76.6\% $\rightarrow$ 29.9\%, $WER^{SUN}_{Test}$: 72.7\% $\rightarrow$ 28.6\%) and closes the gap in performance between PHOENIX-2014 and Scene-PHOENIX significantly.

\noindent\textbf{Evaluation on VAC.} We evaluate the VAC model under same condition as the Baseline experiments. When both BR and DAE are applied to VAC, Ours significantly improves the performance on Scene-PHOENIX, even with only 10 scene images in total (\eg, $K=1$).
\input{tab/abl_backbone}

Moreover, we find that the DAE not only improves the performance of our model on Scene-PHEONIX, but also achieves better performances in WER of Dev and Test splits in PHOENIX-2014.
This indicates that the disentanglement procedure in DAE with BR not only improves the model's robustness to background shifts but also improves the performance of sign videos in monochromatic backgrounds.

\noindent \textbf{Variation on scene images $K$ for training.} By gradually increasing the value of $K$, we also observe that the WER reductions in Scene-PHOENIX are consistent, and DAE indirectly helps BR.
We also highlight that Ours has superior performance to the oracle when $K=1000$ in all metrics.
While the oracle has access to more than 1$M$ images during training, our BR + DAE approach is $1000\times$ more efficient in terms of the number of scene images required, even without using off-the-shelf human segmentation masks.

\subsection{Additional Analysis}
\noindent \textbf{Ablation on additional training data.}
We systematically study the WER scores by ablating additional costs such as pretraining, using pose data, and BR for CSLR model training in \Cref{tab:abl_br}. 
Note that we fix the feature extractor as ResNet-18, and $K=10$ for BR. 
First, pose supervision with extra annotations~\cite{zhou2020spatial} contributes to both WER and $\text{WER}^{\text{SUN}}$ significantly.
Moreover, pretraining feature extractor on ImageNet (denoted as pretrain), while using pose supervision can further reduce the WER scores in Test split: (WER: 24.0\% and $\text{WER}^{\text{SUN}}$: 46.6\%), which shows the best performance throughout the experiments without using extra scene images.
A similar observation can be made when we additionally apply BR. 
While the model trained with BR, pretrain, and pose can greatly improve the performance: (WER: 23.9\% and $\text{WER}^{\text{SUN}}$: 29.4\%), 
the model trained with BR + DAE without pose supervision outperforms it with 23.2\% and 28.6\% WER and $\text{WER}^{\text{SUN}}$ in Test split, respectively. 
We conclude that DAE along with BR outperforms in all metrics with annotation-efficiently, compared to training CSLR models with pose supervision.

\noindent \textbf{Different backbones.}
To show that our proposed framework of BR and DAE are generalizable, 
we show the results obtained by using different feature extractors in \Cref{tab:abl_backbone}. While GoogLeNet~\cite{szegedy2015going} shows slightly worse performance in WER compared to ResNet-18, it shows greater improvements in $\text{WER}^{\text{SUN}}$ 
when our BR + DAE is applied. 

\subsection{Qualitative Results}

\begin{figure}[!h]
    \centering
    \includegraphics[width=.55\linewidth]{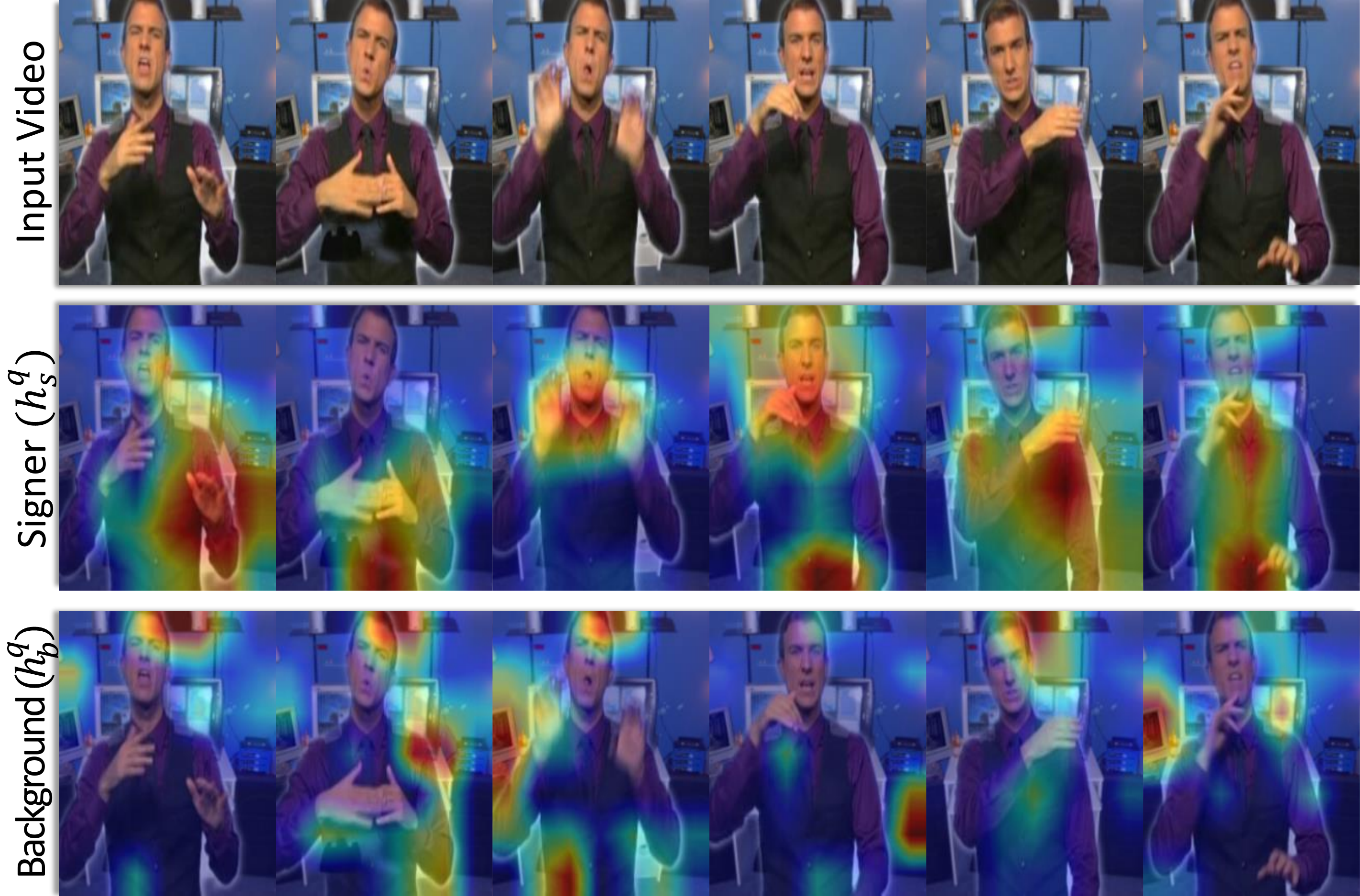}
    \caption{Grad-CAM~\cite{selvaraju2017grad} comparison of the signer feature $h^q_s$ and background feature $h^q_b$ in our framework. 
    By virtue of our Disentangling Auto-Encoder, $h^q_s$ and $h^q_b$ in latent space consistently focus on the signer and background area respectively.
    Verifying that the two components are disentangled in latent space from the background randomized video. 
    }
    \label{fig:DAE_disentanglement}
    \vspace{-2mm}
\end{figure}
\noindent \textbf{Latent Feature Disentanglement.} 
To validate the DAE's ability to distinguish between the signer and background, 
we visualize the Grad-CAM~\cite{selvaraju2017grad} of $h^q_s$ and $h^q_b$ in~\cref{fig:DAE_disentanglement}. We see that the model is able to focus on the regions that make up the signer's features $h^q_s$ (\eg,~hands and face) while the background feature $h^q_b$ focuses on the background region (\ie, the region outside the signer).

\begin{figure}[!h]
    \centering
    \includegraphics[width=.95\linewidth]{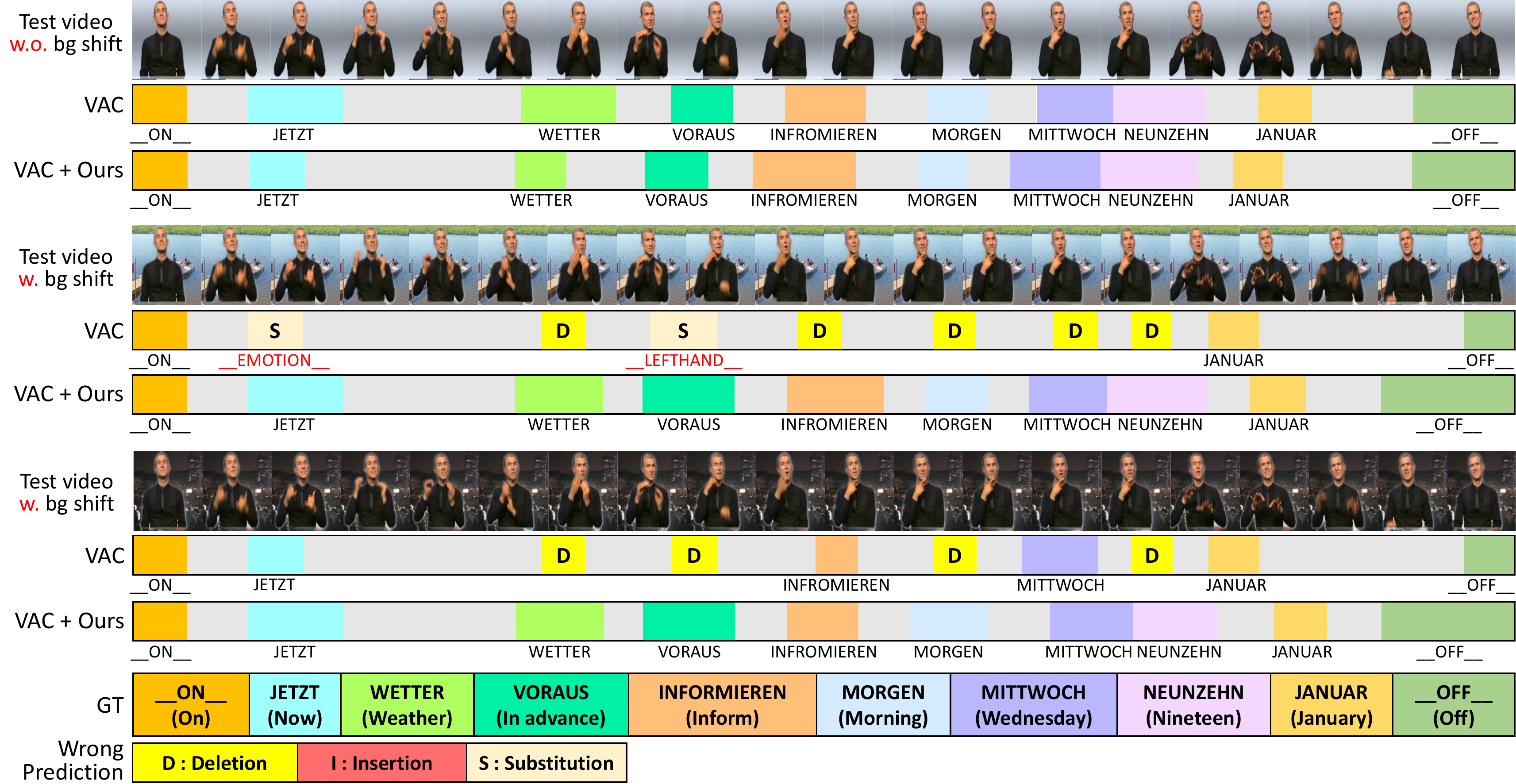}
    \caption{Comprehensive comparison of gloss predictions between VAC~\cite{min2021visual} and Ours. We visualize the frame-level gloss predictions from the models and show the difference when the background shifted. We observe that VAC fails to predict correct glosses with different backgrounds, while our method consistently recognizes glosses regardless of backgrounds.}
    \label{fig:qualitative_pred}
    \vspace{-2mm}
\end{figure}
\noindent \textbf{Gloss Prediction.} 
We qualitatively show our method's robustness to background shifts in~\cref{fig:qualitative_pred}.
we visualize the predicted glosses from three sign language videos, which have the ground truth with different backgrounds. Without background shift, the original VAC~\cite{min2021visual} correctly predicts gloss sequence while it fails when the background shifts. In contrast, Ours correctly predicts gloss sequences in all test videos regardless of background.

%% file: tab/main_table.tex
\begin{table*}[!t]
\centering
\resizebox{0.90\linewidth}{!}{%
    \begin{tabular}{lccccccc}
        \toprule
         &  & \multicolumn{2}{c}{PHOENIX-2014} & \multicolumn{4}{c}{Scene-PHOENIX} \\
         &  & \multicolumn{2}{c}{WER} & \multicolumn{2}{c}{$\text{WER}^{\text{LSUN}}$} & \multicolumn{2}{c}{$\text{WER}^{\text{SUN}}$} \\
        {Method} & $K$ & \multicolumn{1}{c}{Dev} & \multicolumn{1}{c}{Test} & \multicolumn{1}{c}{Dev} & \multicolumn{1}{c}{Test} & \multicolumn{1}{c}{Dev} & \multicolumn{1}{c}{Test} \\
        \midrule
        VAC-Oracle~\cite{min2021visual} & 0.1M+ & 21.5 & 22.0 & 24.3 & 24.2 & 23.8 & 24.1 \\
        \midrule
        Baseline & - & 31.2 & 33.2 & 101.1 & 101.0 & 100.9 & 101.2 \\
        ~~w/ pretrain & - & 25.4 & 26.1                     & 71.0 & 76.6 & 69.9 & 72.7 \\
        ~~\textbf{w/ BR + DAE (Ours)}   & 10 & 23.1 & 23.2  & 30.0 & 29.9 & 27.8 & 28.6 \\
        \midrule
        VAC & - & 21.2 & 22.3 & 65.0 & 68.8 & 66.7 & 67.5  \\
        ~~w/ BR & 1 & 21.9 & 22.9       & 30.0 & 32.4 & 30.5 & 30.5 \\
        ~~w/ BR & 10 & 21.2 & 22.4      & 30.1 & 32.0 & 29.5 & 30.4 \\
        ~~w/ BR & 100 & 21.5 & 21.8     & 30.0 & 31.9 & 31.7 & 30.7 \\
        ~~w/ BR & 1000 & 22.4 & 22.9    & 27.7 & 29.2 & 28.5 & 28.6 \\
        \midrule
        ~~\textbf{w/ BR + DAE (Ours)} & 1    & \textbf{20.6} & \textbf{21.5} & 26.4 & 27.7 & 26.3 & 26.1 \\
        ~~\textbf{w/ BR + DAE (Ours)} & 10   & 20.9 & 21.5 & 26.7 & 27.4 & 26.4 & 26.1 \\
        ~~\textbf{w/ BR + DAE (Ours)} & 100  & 21.5 & 21.9 & 23.7 & 24.0 & 23.3 & 23.6 \\
        ~~\textbf{w/ BR + DAE (Ours)} & 1000 & 20.8 & 21.7 & \textbf{22.9} & \textbf{23.4} & \textbf{22.5}  & \textbf{23.1} \\
        \bottomrule
    \end{tabular}%
}
\caption{Experimental results on PHOENIX-2014 and Scene-PHOENIX. 
VAC-Oracle is a VAC model that is trained on all LSUN background matted images. While the performance of the baselines severely degrades under Scene-PHOENIX, the proposed Background Randomization (BR) shows significant performance improvements. Our final model (BR + DAE) shows the best performance among the baseline models. 
Note that our final model with $K=1$ outperforms all VAC w/ BR models. Moreover, Our with $K=1000$ surpasses the VAC-Oracle and VAC in both dataset without any off-the-shelf human segmentation masks.
} 
\label{tab:main_table}
\end{table*}

%% file: tab/abl_br.tex
\begin{table}[h]
\centering
    \resizebox{0.55\linewidth}{!}{%
        \begin{tabular}{ccccccc}
            \toprule
              &  &  & \multicolumn{2}{c}{WER} & \multicolumn{2}{c}{$\text{WER}^{\text{SUN}}$} \\
              pretrain & Pose & BR & Dev & Test & Dev & Test \\
            \midrule
                           &   &  & 31.2 & 33.2 & 100.9 & 101.2 \\
                          & \cmark  &  & 28.4 & 27.9 & 73.4 & 73.5 \\
                         \cmark & \cmark &  & 23.6 & 24.0 & 51.4 & 46.6 \\
            \midrule
                            &   & \cmark & 29.9 & 31.0 & 48.0 & 48.3 \\
                           & \cmark  & \cmark & 29.0 & 29.1 & 35.1 & 34.8 \\
                          \cmark & \cmark & \cmark & 23.8 & 23.9 & 29.5 & 29.4 \\
            \midrule
            \multicolumn{3}{l}{\bf ResNet-18 w/ BR + DAE} & 23.1 & 23.2 & 27.8 & 28.6  \\
            \bottomrule
        \end{tabular}%
    }
    \caption{Ablation on additional training data. Using DAE is more efficient in annotation cost compared to using pose, which requires extra annotation. We emphasize using additional 100 scene images for BR is much cheaper than annotating pose for training.}
    \label{tab:abl_br}
\end{table}

%% file: tab/abl_backbone.tex
\begin{table}[t]
\centering
    \resizebox{0.46\linewidth}{!}{%
        \begin{tabular}{lcccc}
            \toprule
             & \multicolumn{2}{c}{WER} & \multicolumn{2}{c}{$\text{WER}^{\text{SUN}}$} \\
            Backbone & Dev & Test & Dev & Test \\
            \midrule
            GoogLeNet~\cite{szegedy2015going}   & 26.2 & 27.9 & 76.0 & 80.0 \\
            ~~\textbf{w/ BR + DAE}              & 24.4 & 26.9 & 27.8 & 27.9 \\
            \midrule
            ResNet18~\cite{he2016deep}          & 25.4 & 26.1 & 69.9 & 72.7 \\
            ~~\textbf{w/ BR + DAE}              & 23.1 & 23.2 & 27.8 & 28.6 \\
            \bottomrule
        \end{tabular}%
    }
    \caption{Comparison of performances with different feature extractors: GoogLeNet~\cite{szegedy2015going} and ResNet18~\cite{he2016deep}. Our framework consistently works well with different feature extractors.
    }
    \label{tab:abl_backbone}
\end{table}

%% file: sec/5_conclusions.tex
\section{Conclusion}
In this paper, we introduce a new Scene-PHOENIX benchmark, composed of background-synthesized sign videos, and a pipeline for automatically generating CSLR benchmark with backgrounds to expand the possibilities for real world applications for CSLR.
We also propose a simple but effective Background Randomization (BR) that shows dramatic performance gains without human segmentation masks and Disentangling Auto-Encoder (DAE) that disentangles the signer features from the background in latent space.
Experiments on two datasets (PHOENIX-2014, Scene-PHOENIX) demonstrate strong robustness to background shifts while maintaining the existing model performance.

%% file: sec/6_ack.tex
\section{Acknowledgement}
This work was supported by Institute of Information \& communications Technology Planning \& Evaluation (IITP) grant funded by the Korea government (MSIT, No. 2022-0-00989, Development of Artificial Intelligence Technology for Multi-speaker Dialog Modeling).

%% file: sec/6_supplementary.tex
We present additional quantitative and qualitative results that could not be included in the main paper due to space limitations. All references, figures, and tables in this supplementary material are self-contained.

\section{Automatic Benchmark Generation Pipeline}
\Cref{alg:benchmark} describes the pipeline with psuedo-code for generating background robustness benchmark dataset, named Scene-PHOENIX, utilizing existing CSLR benchmark data and common scene images.
Note that we also include the actual self-contained codes as supplementary that reproduce our Scene-PHOENIX benchmark dataset used in the experiments, and that we will release the codes to public after the review period. 

\begin{figure*}[t]
    \centering
    \includegraphics[width=.7\linewidth]{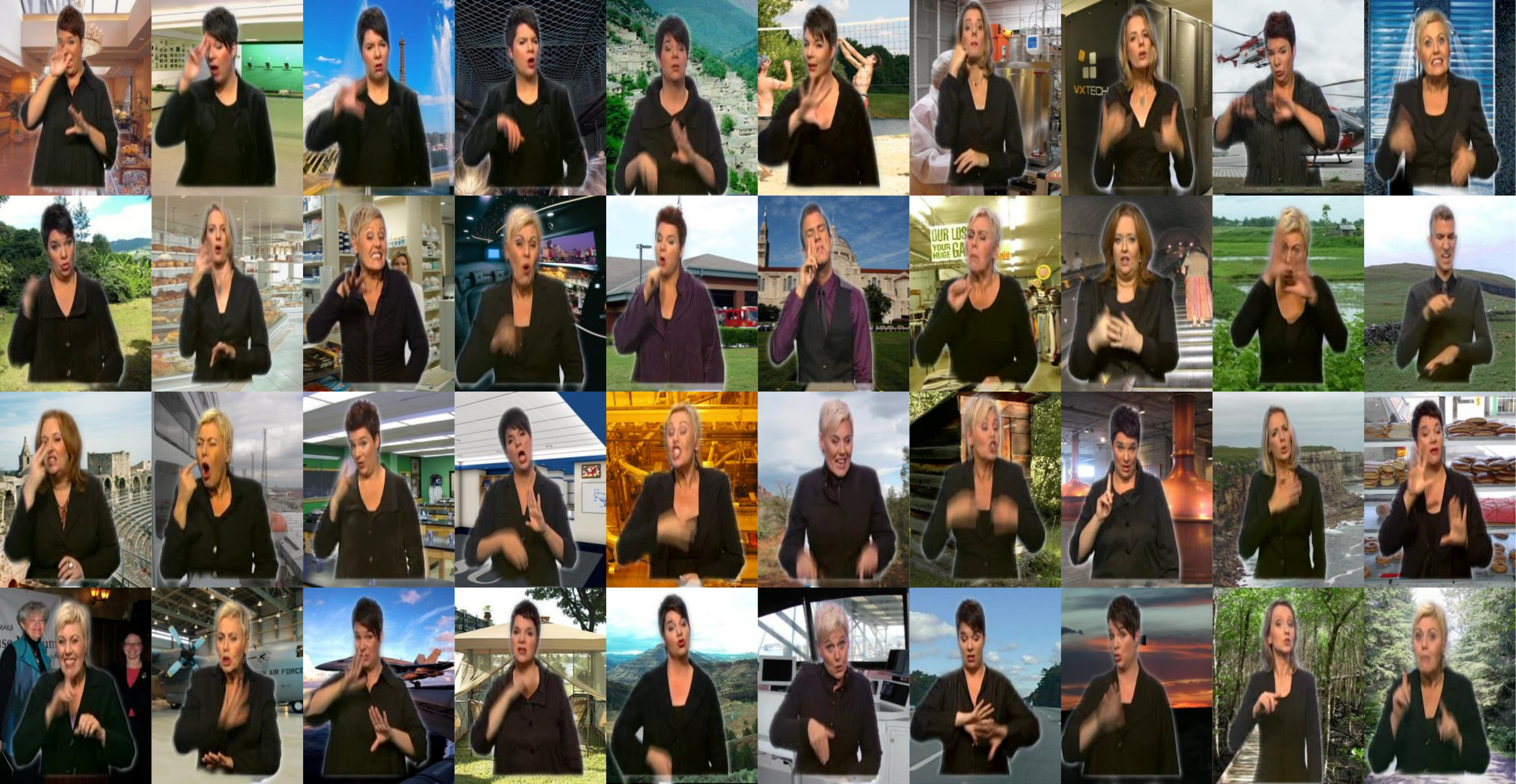}
    \caption{Examples of video samples with backgrounds for evaluating the background robustness of the model. To best our knowledge, we are the first to construct a benchmark dataset with various backgrounds that can evaluate the background robustness of the CSLR model, and it has a reasonable construction cost in that it reuses existing CSLR datasets and scene datasets.}
    \label{fig:test_data}
\end{figure*}

In line 3: we construct a subset of the scene data so that each scene class of scene data is uniformly distributed across the CSLR data. Here, the number of images in the subset is identical to the number of videos in CSLR data.
In line 5: we randomly assign a scene class to the video, and then a scene image of that class is further selected from the subset.  
These points are also clarified in Sec. {\color{red} 3.2} in the main paper.
\cref{fig:test_data} visualizes the \emph{randomly selected} test data of Scene-PHOENIX, generated from the proposed pipeline. 

\section{Training Data Generation Process}
\begin{figure*}[!t]
    \centering
    \includegraphics[width=.9\linewidth]{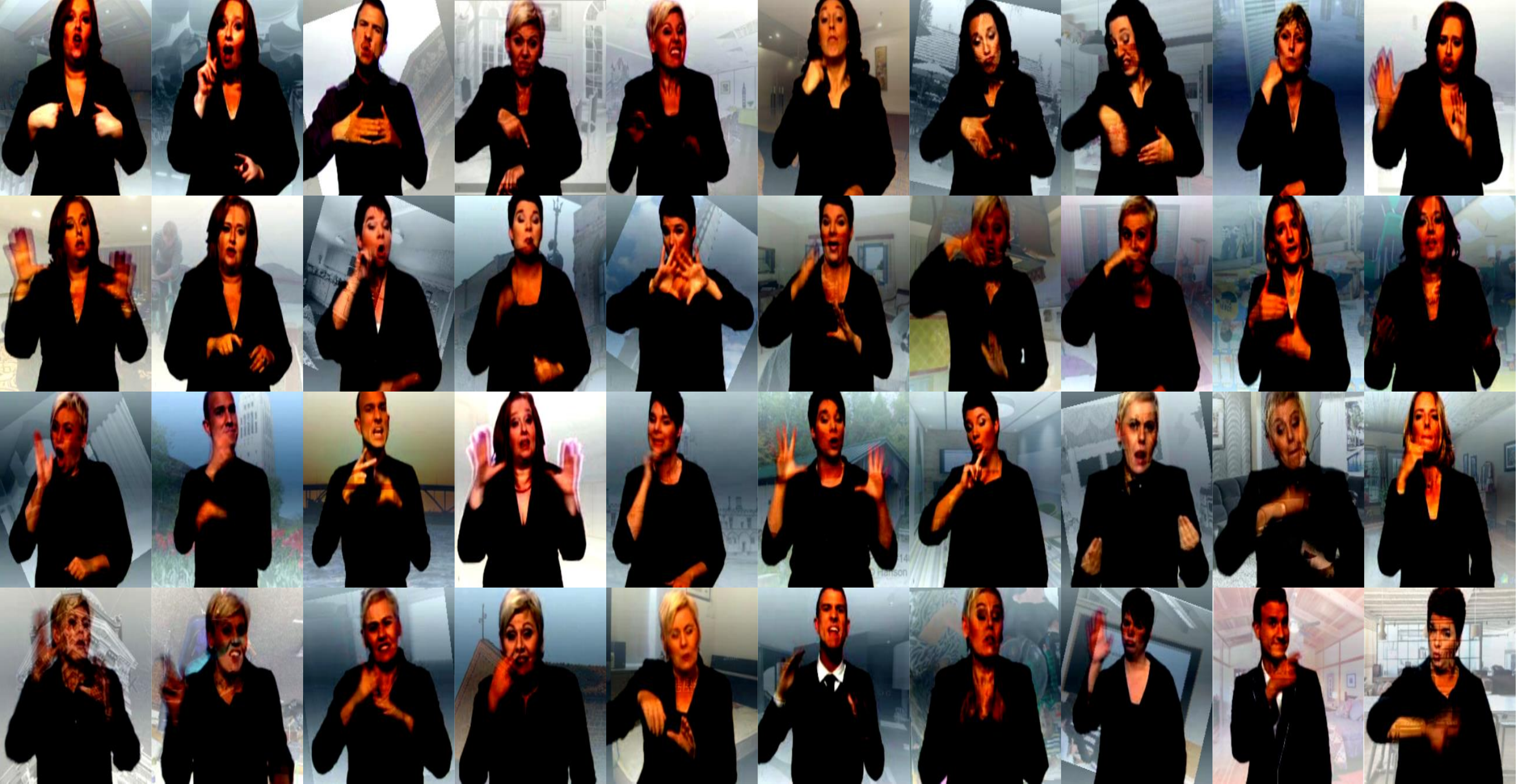}
    \caption{Examples of video samples with backgrounds used in the learning process. As augmentation for background only, color jitter and random rotation are applied for the background augmentation. We emphasize that it is possible to train the robustness of CSLR models to the background without using expensive human masks during the training process.}
    \label{fig:train_data}
\end{figure*}

As explained in the main paper, we adapt mix-up~\cite{zhang2017mixup} method to our framework to generate background randomized training data. With the mix-up, our Background Randomization (BR) can be formulated as follow:

\begin{align} \label{eq:mix_up}
    I = \lambda{I_{back}} + (1-\lambda)I_{sign}.
\end{align}
$I$ denotes a frame of background randomized sign video. $I_{back}$ and $I_{sign}$ are a background image randomly selected from LSUN dataset and a frame of original sign video respectively. Here, $\lambda$ is a weight factor between a background image and a sign frame, and $\lambda$ is randomly determined within the range of 0.1 to 0.6. Additionally, 
when employing Background Randomization (BR), we apply color jitter to the background scene image to augment limited background image data. As note, \cref{fig:train_data} illustrates the examples after applying BR.

\begin{algorithm}[t]
    \centering
    \caption{Benchmark dataset generation}
    \label{alg:benchmark}
    \begin{algorithmic}[1]
        {\small
        \STATE {\bfseries Require:} CSLR data, scene data, person segmentation model
        \STATE {\bfseries Initialize:} $\cV \leftarrow \emptyset$ \vspace{1mm}
        \STATE {Select} scene images {from} scene data.
        \FOR{{\tt video} {\bfseries in} CSLR data}
            \STATE Assign a {\tt scene} image to {\tt video}.
            \STATE Run person segmentation model on {\tt video} to get {\tt mask}.
            \STATE ~
            \STATE \COMMENT{Fig. {\color{red} 3(a)} in the main paper.}
            \STATE $\cV \leftarrow \cV \; \cup$ \textit{Matting}({\tt video}, {\tt scene}, {\tt mask})
        \ENDFOR \vspace{1mm}
        \STATE {\bfseries Output:} CSLR data with background scenes $\cV$
        }
    \end{algorithmic}
\end{algorithm}

\section{Training Details}
\subsection{Dataset.} 
We utilize publicly available PHOENIX-2014~\cite{koller2015continuous} dataset which has been widely adopted in CSLR to verify the claims made in this work and validate our framework.
This dataset contains 6841 videos with 9 different signers with a vocabulary size of 1232, where the number of videos for Train, Dev and Test are 5672, 540, and 629, respectively.
The videos are crawled from weather forecasts and have a monochromatic background.

\subsection{Implementation details}
The proposed Disentangling Auto-Encoder (DAE) has encoder-decoder architecture, and both the encoder and the decoder consist of two fully-connected layers.

The video frames are first resized to 256$\times$256, followed by random crop of 224$\times$224 size at the same location of all the frames and random horizontal flip with 50\% probability. We then randomly insert duplicated frames up to 20\% in total length, followed by random deletion of frames up to 20\% of the whole length. 
We train the network using Adam optimizer~\cite{kingma2014adam} with batch size 2 and weight decay $10^{-4}$ for 100 epochs. 
The initial learning rate is $10^{-4}$ and the cosine scheduler is adopted to decay the rate.
The final objective of Ours when integrated with VAC~\cite{min2021visual} network is as follows:
\begin{equation}
    L_{\text{total}} = \underbrace{L_{\text{CTC}} + L_{\text{VE}} + \alpha L_{\text{VA}}}_{L_{\text{VAC}}} + \underbrace{L_{\text{sim}} + L_{\text{rec}}}_{L_{\text{DAE}}},
\end{equation}
where the first three terms correspond to VAC, and the final two terms are used for train our DAE. 
In our framework, we empirically find that lower value of $\alpha$ accelerates the convergence of the model as a whole.
Therefore, we change $\alpha$ value from 25 to 5. Our framework is implemented using PyTorch~\cite{paszke2019pytorch}.

\subsection{Evaluation Metric.}
To evaluate the performance of CSLR models, Word Error Rate (WER)~\cite{koller2015continuous} is widely adopted.
It first counts the number of operations of ``substitution,'' ``deletion,'' and ``insertion'' required to align the recognized gloss sequence with the reference, then, the total number of operations is divided by the length of the reference:
\begin{equation}
    \text{WER} = \frac{\text{\#substitutions} + \text{\#deletions} + \text{\#insertions}}{\text{\#words in reference}},
\end{equation}

\section{Ablation Study}
We further dissect the design of the proposed framework including Background Randomization (BR) and Disentangling Auto-Encoder (DAE) to investigate each component's contribution to the overall robustness in background shift. 
As note, the number of images $K$ per scene class for BR (explained in the main paper) is set to 10.

\subsection{Component Analysis on BR.}
\Cref{tab:sup_br} ablates different design choices for the sign video with randomized background, while training our framework along with VAC~\cite{min2021visual}.
First, 
we generate the input video corresponding to the \emph{query} encoder by just applying color jitter in~\cite{he2020momentum} to the monochromatic sign video without any scene images.
This shows moderate performances on PHOENIX-2014, but performs poorly on Scene-PHOENIX, which calls for the realistic scene images for simulating background shift. 
As such, when we randomize the background during training without human masks, robustness can be improved significantly ($\text{WER}^{\text{SUN}}$ Test: 48.1\% $\rightarrow$ 28.1\%).
In addition, further applying color jitter to the background scene image obtains the best results.

\input{tab/sup_abl_br}
\subsection{Similarity Loss Analysis.}
The similarity loss $L^{neg}_{\text{sim}}$ between the background pairs $h^q_b$ and $h^k_b$ more penalizes visually similar background pairs. Here, we evaluate the different margin value $\Delta$ in the loss $L^{neg}_{\text{sim}}$, as shown in~\Cref{tab:sup_margin}.
When the margin is set to 0.25, the difference between WER and $\text{WER}^{\text{SUN}}$ values is smaller than when the margin is 0. 
Finally, our model performs best when the margin is 0.5.
This suggests that when the similarity between the background pairs $h^q_b$ and $h^k_b$ is high because of small $\lambda$ in~\Cref{eq:mix_up}, rather attenuating the penalty is beneficial for the robustness.

\input{tab/sup_margin}

\subsection{Component Analysis on DAE.}
In~\Cref{tab:sup_component}, we validate two losses constituting Disentangling Auto-Encoder (DAE) on the robustness in background shift: the similarity loss $L_\text{sim}$ in latent space and the reconstruction loss $L_{\text{rec}}$ after the decoder. 
When only $L_{\text{rec}}$ is employed without $L_\text{sim}$, the performance gap between WER and $\text{WER}^{\text{SUN}}$ corresponds to 6.4\% and 6.1\% in the Dev and Test split, respectively.
When only $L_\text{sim}$ is used, the performance gap can be reduced to 5.5\% and 5.3\% in the Dev and Test split.
This demonstrates both $L_\text{rec}$ and $L_\text{sim}$ are essential to improve robustness.
Moreover, we obtain the best results in all the metrics when both of $L_\text{rec}$ and $L_\text{sim}$ are applied.

\input{tab/sup_component}
\section{Qualitative results}

\subsection{Gloss Prediction.}
\begin{figure*}[t]
    \centering
    \includegraphics[width=.90\linewidth]{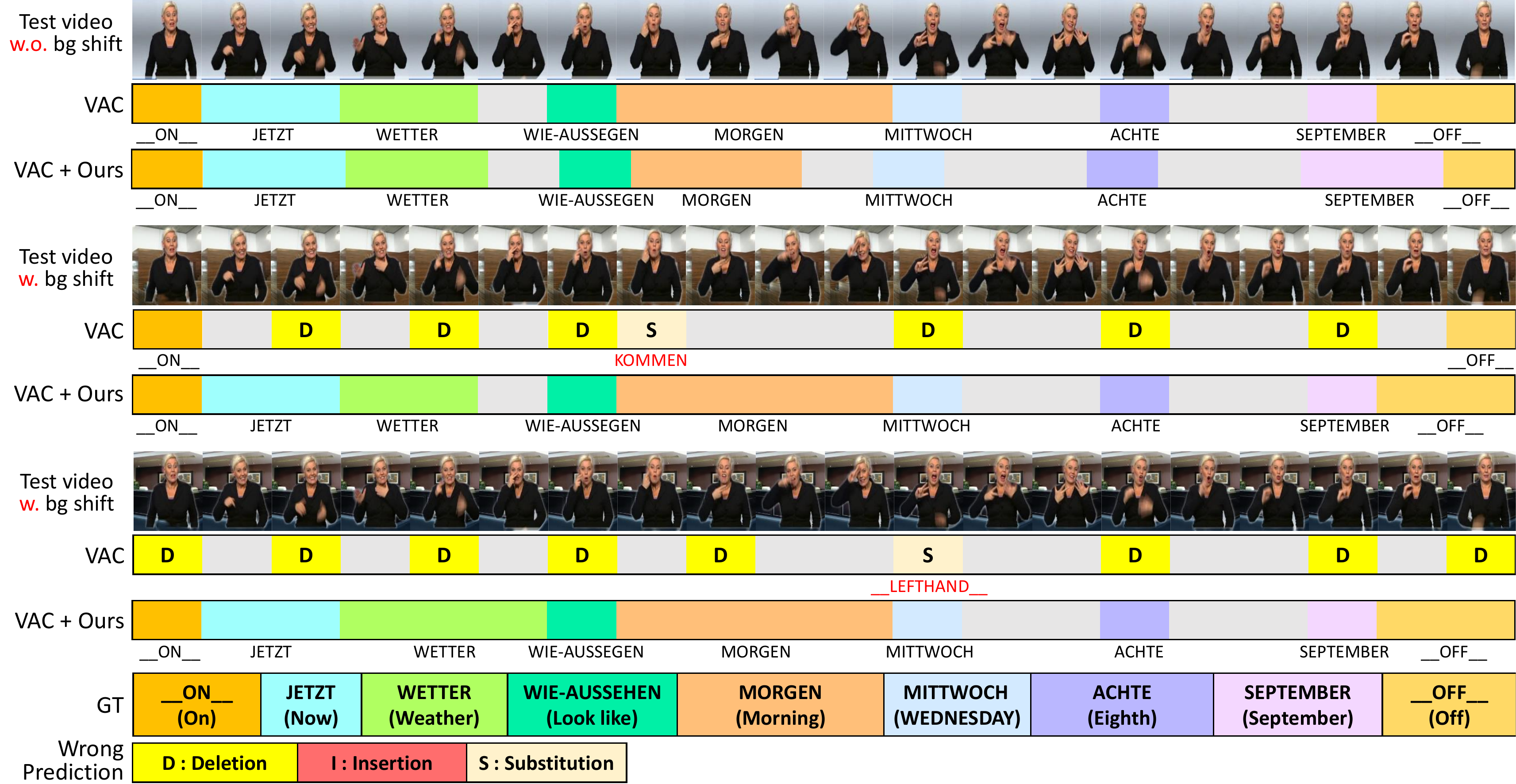}
    \caption{Comprehensive comparison of gloss predictions between VAC~\cite{min2021visual} and Ours. We visualize the frame-level gloss predictions
    from the models and show the difference when background is changed. We note this case as \emph{Case 1} when both VAC and Ours exactly recognition whole glosses from the input video with monochromatic background. In this case, only a few glosses predicted by VAC are correct, when the background of video changes. while Ours consistently recognizes glosses regardless of the background change.}
    \label{fig:qualitative_1}
\end{figure*}

\begin{figure*}[!t]
    \centering
    \includegraphics[width=.90\linewidth]{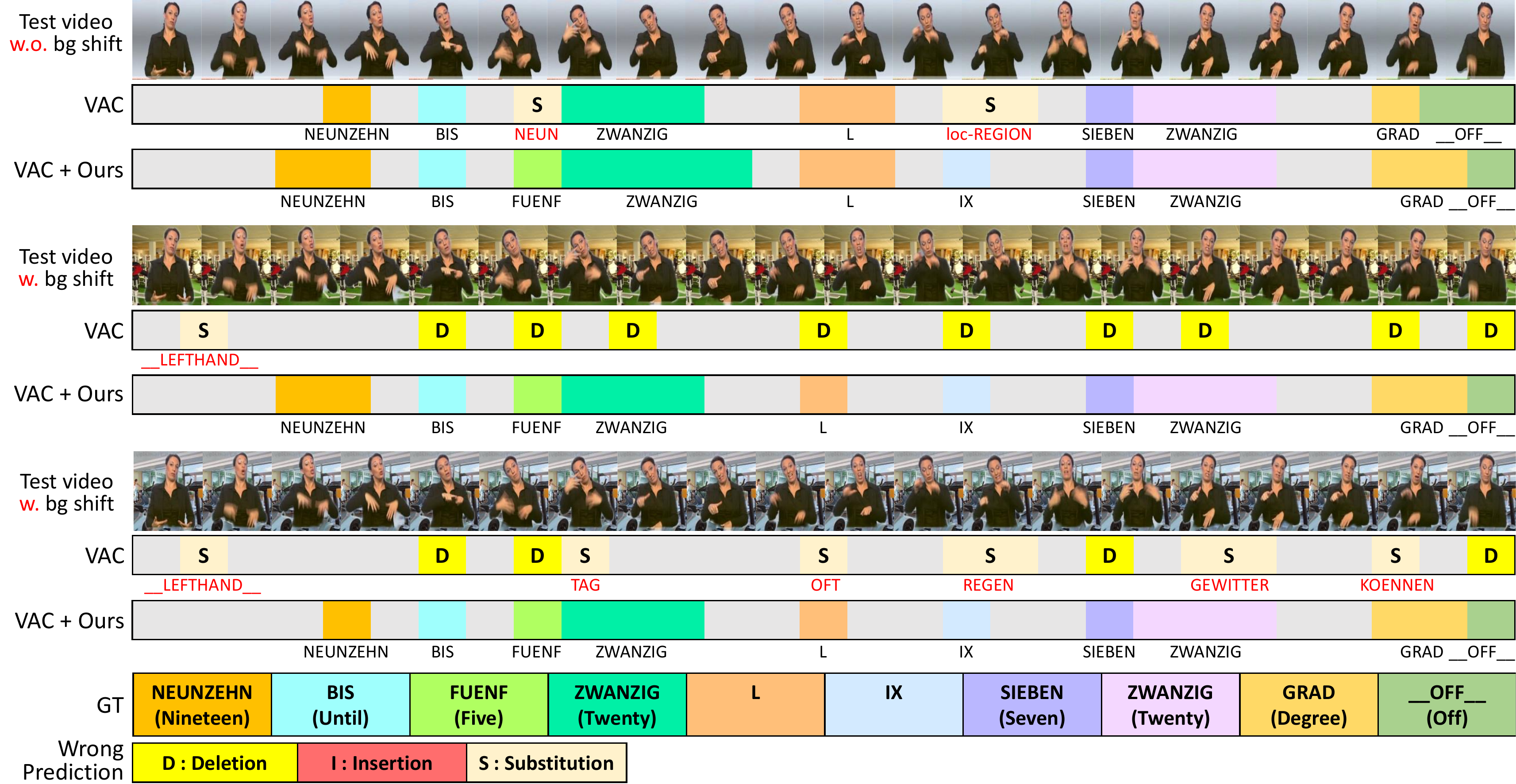}
    \caption{Comprehensive comparison of gloss predictions between VAC~\cite{min2021visual} and Ours. We note this case as \emph{Case 2} when only Ours correctly predict the whole gloss sequence from the input video with monochromatic background. In this case, VAC totally output wrong gloss sequence, when the background of video changes. while Ours consistently recognizes glosses regardless of the background change. From this, we emphasize our method improves not only the gloss recognition performance but also the robustness to background change.}
    \label{fig:qualitative_2}
\end{figure*}

We visualize more qualitative results of our framework. \cref{fig:qualitative_1} is \emph{Case 1} when VAC~\cite{min2021visual} and Ours correctly perform well to recognize all glosses composing one sentence from a given sign video with monochromatic background. The output of the VAC is greatly corrupted when the background of the input video changes, failing to accurately predict the gloss sequence. On the other hand, Ours consistently shows almost no change in the output sequence even when the background of the input video totally changes. Interestingly, when the VAC fails to predict the correct gloss sequence, the insertion phenomenon that occurs when predicting words longer than the sentence length does not occur. \cref{fig:qualitative_2} is \emph{Case 2} when VAC fails to predict the gloss sequence and ours succeeds from an original video without background change. In this case, the output of the VAC differs from the ground truth more severely when compared to \emph{Case 1}. Even in this case, the insertion phenomenon does not occur as in \emph{Case 1}. On the other hand, Ours consistently shows the correct answer from two videos having different backgrounds.

\subsection{Latent Feature Disentanglement.}
\begin{figure}[t]
    \centering
    \includegraphics[width=.50\linewidth]{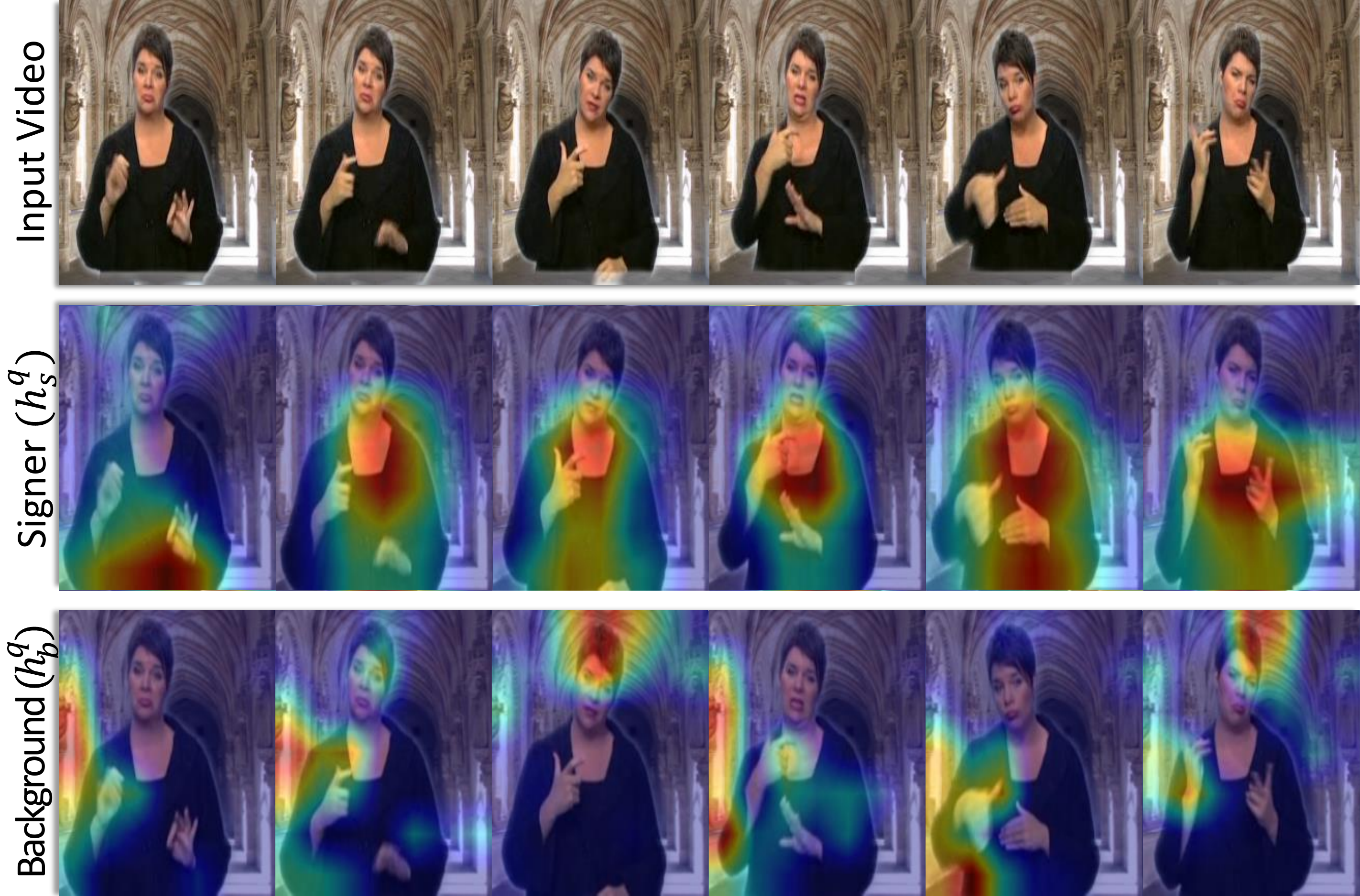}
    \caption{Feature activation visualization of both signer and background latent features via Grad-CAM. Our Disentangling Auto-Encoder (DAE) consistently focus on the signer and background regions respectively during the feature reconstruction process. From this, we demonstrate our DAE can disentangle the latent features into background and signer in latent space.}
    \label{fig:DAE}
\end{figure}

We provide more visual examples about the disentangled latent feature representations via Grad-CAM~\cite{selvaraju2017grad} in~\cref{fig:DAE}.
As we explained in main paper, $h_s^q$ and $h_b^q$ denote the latent signer and background features from query encoder respectively. Our framework is able to capture the important areas for reconstructing the original features from the latent features. From signer's latent features, $h_s^q$ reconstructs the origin features focusing on the signer regions. On the other hand, $h_b^q$ rebuild the origin features focusing on the background region (\ie, the region outside the signer). We emphasize that DAE does not require any extra annotations or off-the-shelf detector for physically dividing the latent features into the background region and the signer region.

\section{Additional Discussion}

\noindent \textbf{Potential Societal Impact.}
Our proposed method contributes directly to AI model's translation of sign language recognition, which is an important task for the betterment of society. As mentioned throughout the paper, the limitations of the current dataset calls for the need of our work so that robust CSLR models can be deployed in society.

\noindent \textbf{Limitation and Future Work.}
Although we first point out the background shift problem in CSLR field, the proposed Scene-PHOENIX benchmark still have some differences with real world environments. In order to break down the language barrier between deaf people and non-deaf people, new CSLR dataset with more realistic environment (\Eg, moving backgrounds, different camera views, various signers, \etc.) might be necessary. We suggest that future research in CSLR should include the study of robustness of models outside the studio to truly facilitate communication between the hearing and hard-of-hearing people in real life.
\clearpage

%% file: tab/sup_abl_br.tex
\begin{table}[t]
\vspace{2mm}
\centering
    \resizebox{0.45\linewidth}{!}{%
        \begin{tabular}{cccccc}
            \toprule
            &  & \multicolumn{2}{c}{WER} & \multicolumn{2}{c}{$\text{WER}^{\text{SUN}}$} \\
            Jitter      & Scene     & Dev  & Test & Dev  & Test \\
            \midrule
            \cmark      &           & 21.3 & 22.0 & 51.5 & 48.1 \\
                        & \cmark    & 21.1 & 21.9 & 27.5 & 28.1 \\
            \cmark      & \cmark    & \bf 20.9 & \bf 21.5 & \bf 26.4 & \bf 26.1 \\
            \bottomrule
        \end{tabular}%
    }
    \caption{Ablation study on the different design choices for the background randomization (BR). 
    To improve the robustness, employing additional scene images (Scene) while applying color jitter to the scene image (Jitter) is the most effective.
    }
    \label{tab:sup_br}
    \vspace{-2mm}
\end{table}

%% file: tab/sup_margin.tex
\begin{table}[t]
\centering
    \resizebox{0.45\linewidth}{!}{%
        \begin{tabular}{ccccc}
            \toprule
             & \multicolumn{2}{c}{WER} & \multicolumn{2}{c}{$\text{WER}^{\text{SUN}}$} \\
            Margin ($\Delta$) & Dev & Test & Dev & Test \\
            \midrule
            0.0   & 21.2 & 22.4       & 29.1 & 29.2 \\
            0.25  & 22.1 & 22.4       & 27.3 & 27.6 \\
            0.5   & \bf 20.9 & \bf 21.5       & \bf 26.4 & \bf 26.7 \\
            \bottomrule
        \end{tabular}%
    }
    \caption{Ablation study on different values of $\Delta$. 
    If the margin is 0, the similarity loss falls to the correlation score itself.
    If the margin is larger than 0, the model does not penalize the background pair with a correlation score less than the margin.}
    \label{tab:sup_margin}
\end{table}

%% file: tab/sup_component.tex
\begin{table}[t]
\centering
    \resizebox{0.45\linewidth}{!}{%
        \begin{tabular}{cccccc}
            \toprule
              &  & \multicolumn{2}{c}{WER} & \multicolumn{2}{c}{$\text{WER}^{\text{SUN}}$} \\
              $L_\text{rec}$ & $L_\text{sim}$ & Dev & Test & Dev & Test \\
            \midrule
                &  & 21.2 & 22.4 & 29.5 & 30.4 \\
                \cmark &  & 21.6 & 22.2 & 28.0 & 28.3 \\
                  & \cmark  & 21.5 & 21.9 & 27.0 & 27.2 \\
                 \cmark & \cmark & \bf 20.9 & \bf 21.5 & \bf 26.4 & \bf 26.1 \\
            \bottomrule
        \end{tabular}%
    }
    \caption{Ablation study on each the component in DAE: reconstruction loss $L_\text{rec}$ and similarity loss $L_\text{sim}$. All the components for training DAE improve the performance, 
    and combining the whole components shows the best performances in all the metrics.
    }
    \label{tab:sup_component}
\end{table}